\documentclass[letterpaper]{article} 

\usepackage{aaai2026}  

\usepackage{times}  
\usepackage{helvet}  
\usepackage{courier}  
\usepackage[hyphens]{url}  
\usepackage{graphicx} 
\usepackage{natbib}  
\usepackage{caption} 
\usepackage{algorithm}
\usepackage{newfloat}
\usepackage{listings}
\DeclareCaptionStyle{ruled}{labelfont=normalfont,labelsep=colon,strut=off} 
\floatstyle{ruled}
\newfloat{listing}{tb}{lst}{}
\floatname{listing}{Listing}

\usepackage{graphicx, url}
\usepackage{amsmath, amssymb, mathtools}
\usepackage{bibunits}
\usepackage{dsfont} 
\usepackage{placeins}

\usepackage{listings}
\lstdefinestyle{mystyle}{
    backgroundcolor=\color{backcolour},   
    commentstyle=\color{codegreen},
    keywordstyle=\color{magenta},
    numberstyle=\tiny\color{codegray},
    stringstyle=\color{codepurple},
    basicstyle=\footnotesize,
    breakatwhitespace=false,         
    breaklines=true,                 
    captionpos=t,                    
    keepspaces=true,                 
    numbers=left,                    
    numbersep=5pt,                  
    showspaces=false,                
    showstringspaces=false,
    showtabs=false,                  
    tabsize=2
}
\usepackage{enumitem}
\usepackage{amsmath, amssymb}
\usepackage{algorithm}
\usepackage{algpseudocode}
\usepackage{booktabs}
\usepackage{multirow}
\usepackage{multicol}
\usepackage{tabularx}
\usepackage{subcaption}
\usepackage{diagbox}

\usepackage[font=footnotesize]{caption}
\newcommand{\highlight}[1]{\textbf{\underline{#1}}}

\usepackage{makecell, tabularx, multirow,  adjustbox}

\usepackage{tcolorbox}
\usepackage[capitalise, nameinlink]{cleveref}

\usepackage{xspace}

\title{\LARGE \bf
Exploring Spatial Representation to Enhance LLM Reasoning in Aerial Vision-Language Navigation
}

\author{
Yunpeng Gao\textsuperscript{\rm 1,2}\equalcontrib, 
Zhigang Wang\textsuperscript{\rm 2}\equalcontrib, 
Pengfei Han\textsuperscript{\rm 1}, 
Linglin Jing\textsuperscript{\rm 2}, 
Dong Wang\textsuperscript{\rm 2}, 
Bin Zhao\textsuperscript{\rm 1,2, \dag}\\
}
\affiliations{
    \textsuperscript{\rm 1}Northwestern Polytechnical University
    \textsuperscript{\rm 2}Shanghai AI Laboratory\\
    gaoyunpeng@mail.nwpu.edu.cn,
    wangzhigang@pjlab.org.cn,
    chenxihan@mail.nwpu.edu.cn,\\
    jing@lboro.ac.uk,
    wangdong@pjlab.org.cn,
    bin@nwpu.edu.cn
}

\begin{document}

\maketitle

\begin{abstract}
Aerial Vision-and-Language Navigation (VLN) is a novel task enabling Unmanned Aerial Vehicles (UAVs) to navigate in outdoor environments through natural language instructions and visual cues.
    However, it remains challenging due to the complex spatial relationships in aerial scenes.
    In this paper, we propose a training-free, zero-shot framework for aerial VLN tasks, where the large language model (LLM) is leveraged as the agent for action prediction.
    Specifically, we develop a novel Semantic-Topo-Metric Representation (STMR) to enhance the spatial reasoning capabilities of LLMs. This is achieved by extracting and projecting instruction-related semantic masks onto a top-down map, which presents spatial and topological information about surrounding landmarks and grows during the navigation process. At each step, a local map centered at the UAV is extracted from the growing top-down map, and transformed into a matrix representation with distance metrics, serving as the text prompt to LLM for action prediction in response to the given instruction.
    Experiments conducted in real and simulation environments have proved the effectiveness and robustness of our method, achieving absolute success rate improvements of 26.8\% and 5.8\% over current state-of-the-art methods on simple and complex navigation tasks, respectively. The dataset and code will be released soon.
\end{abstract}

\section{Introduction}
\label{sec:intro}

The Aerial Vision-and-Language Navigation (Aerial VLN) \cite{liu_2023_AerialVLN} emerges as a groundbreaking task. It enables unmanned aerial vehicles (UAVs) to interpret natural language instructions and visual information to navigate in outdoor environments. This technology can eliminate the necessity for manual UAV operation by human pilots, clearly mitigating the barriers to human-UAV interaction and potentially benefitting rescue, search,  and delivery tasks.

\begin{figure}[htb]
    \centering
    \includegraphics[width=1.0\linewidth]{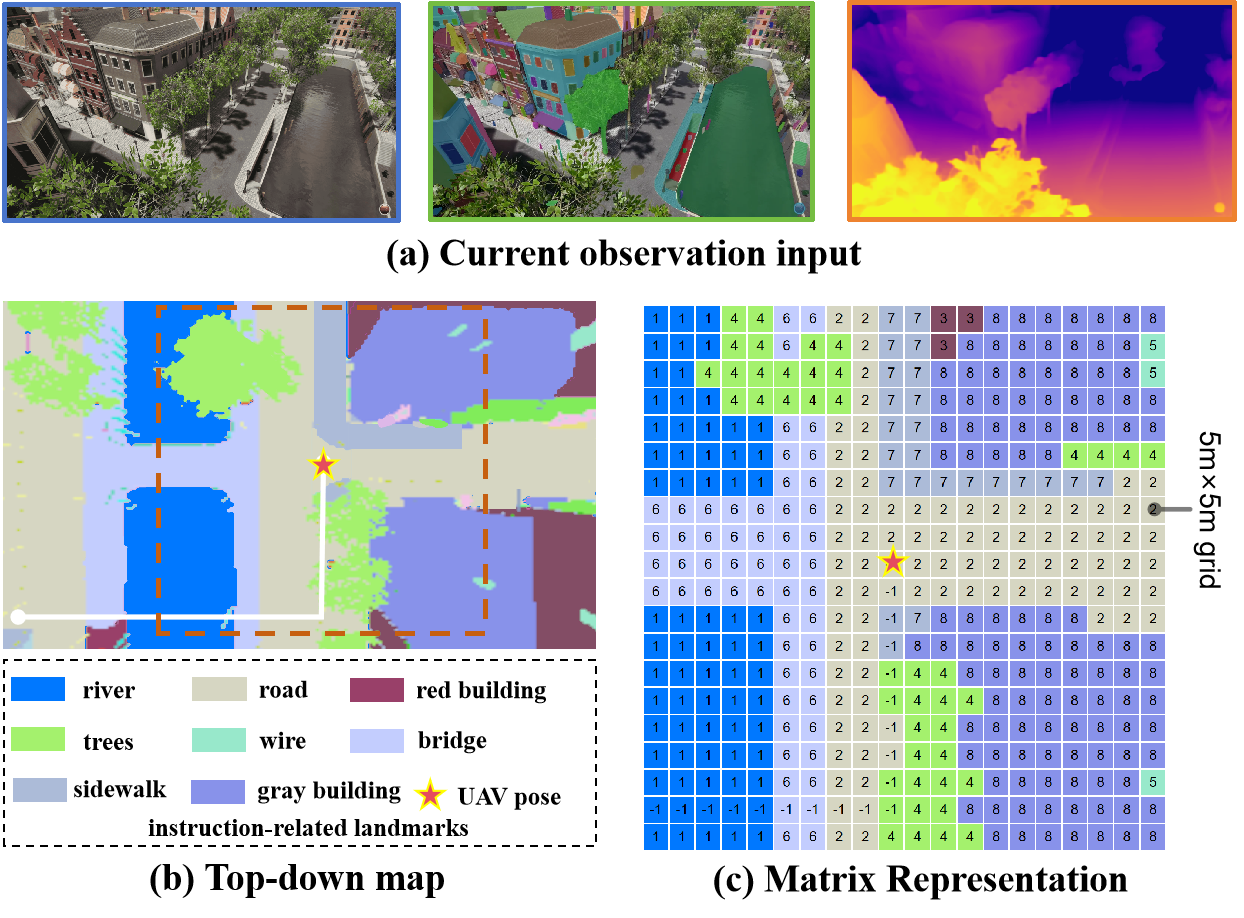}
    \caption{The pipeline to obtain STMR. (a) The observed RGB image, the corresponding segmented image, and the depth image. (b) Segmented images are projected into the top-down map gradually during the UAV flight, which captures the semantic and topological information of the environment. (c) The top-down map is further transformed into a 20x20 matrix representation with distance metrics for LLM reasoning.}
    \label{fig:STMR}
    \vspace{-10pt}
\end{figure}


Recently, VLN tasks have been well-developed. Considering the powerful reasoning capability of large language models (LLMs) and vision language models (VLMs) \cite{ahn2022can, song2023llm, driess2023palm, liu2023visual, liu2024improved}, several VLN methods have started to use LLMs or VLMs as agents to parse instructions and predict actions \cite{zhou2024navgpt, shah2023lm, li2024tina, lin2024navcot}. Specifically, LLM-based methods attempt to describe the visual observations with text to enhance the LLM's scene understanding ability \cite{zhou2024navgpt,chen2024mapgpt}. VLM-based methods integrate the visual and text data and generates context-aware decisions, including~\cite{zhang2024vision, zhang2024sparsevlm, li2024vln, wen2025zero}. 

Although existing LLM-based and VLM-based methods have made significant progress in indoor or ground-based outdoor environments, they still struggle to effectively encode the large-scale spatial information from the aerial view. As shown in ~\cref{fig:STMR}, the aerial scene can be highly complex, which may lead to the overemphasis of instruction-irrelevant objects or the failure to capture the contextual relationships among different areas. Worse still, existing VLMs suffer from limitations in spatial reasoning when taking raw observations as input~\cite{spatialvlm, mindthegap}. Thus, there is an urgent need for the development of VLN methods that effectively integrate both semantic information and precise spatial representations to improve adaptability in aerial scenarios.


To overcome these challenges, we propose a zero-shot LLM-based aerial VLN framework that encodes natural language instructions, RGB images and depth images as text prompt, and generates action predictions (\emph{e.g.,} go forward 10 meters) through LLMs directly. Specifically, a Semantic-Topo-Metric Representation (STMR) is designed for LLMs spatial reasoning. Firstly, the instruction-related landmarks are extracted and corresponding semantic masks are generated via visual perception models, \emph{i.e.}, Grounding DINO \cite{liu2023grounding} and Tokenize Anything \cite{pan2023tap}. Then, the semantic mask is projected into a top-down map as shown in~\cref{fig:STMR} (b). This top-down map encompasses both the UAV's travel trajectory and spatial information, and it grows gradually during the navigation process. To encode the visual information into an LLM-friendly form, we firstly select a fixed-size region centered on the UAV in the top-down map as a local map, then separate the map into equally spaced grids and substitute each grid with a semantic number. As can be seen from \cref{fig:STMR} (c), the matrix representation encompasses topological, semantic, and metric information. It is input to an LLM together with historical actions and text instructions to infer the next action. Experiments have shown that the proposed method significantly enhances the capability of spatial reasoning, and achieves absolute success rate improvements of 26.8\% and 5.8\% over current state-of-the-art methods on simple and complex navigation tasks.

Our contributions are summarized as follows:
\begin{itemize}  
    \item To our knowledge, we design the first LLM-based training-free framework for the aerial VLN task, facilitating the development of UAV navigators. Without an extra action planner, the proposed framework allows for easy integration. 
    \item We propose the Semantic-Topo-Metric Representation (STMR), a unique matrix representation that encompasses topological, semantic, and metric information. STMR is designed to enhance the spatial-aware reasoning capabilities of LLMs in outdoor environments. 
    \item Extensive experiments on the aerial VLN task demonstrate that the proposed method outperforms previous state-of-the-art methods by a large margin, establishing a strong baseline for future zero-shot aerial VLN tasks.
\end{itemize}

\section{Related Work}
\label{sec:related work}

\paragraph{Vision-Language Navigation (VLN).}
VLN aims to enable autonomous agents to navigate complex environments by understanding and executing natural language instructions based on visual context.
Early VLN methods use sequence-to-sequence LSTMs to predict low-level actions~\cite{anderson2018vision} or high-level actions from panoramas~\cite{fried2018speaker}. 
Several attention processes have been proposed~\cite{qi2020object, hong2020language, an2021neighbor} to enhance the process of learning visual textual correspondence. Reinforcement learning is also explored to improve policy learning~\cite{wang2018look, tan2019learning, wang2020soft}. 
Besides, transformer-based architecture have shown superior performance to long-distance contextual information~\cite{hao2020towards, majumdar2020improving}. More recent works~\cite{zhou2024navgpt,chen2024mapgpt} leverage the reasoning and dialogue capabilities of LLMs, achieving great progress.
However, most of them operate in limited spaces on the ground. In contrast, aerial VLN remains challenging due to the large-scale and complex environments.

\paragraph{UAV Navigation.}
Unmanned Aerial Vehicle (UAV) navigation has seen a surge of interest over the past years. Many earlier works~\cite{blukis2018following, blukis2018mapping} using a combination of supervised and imitation learning for efficient training and low-level velocity prediction to guide UAV's control actions. LINGUNET~\cite{misra2018mapping} decomposes instruction execution into two stages, separately using supervised learning for goal prediction and policy gradient for action generation. AerialVLN \cite{liu_2023_AerialVLN} contributes a much more challenging aerial VLN dataset focusing on outdoor aerial environments, and provides a look-ahead guidance method as the baseline. Recent research~\cite{openuav, sautenkov2025uav, tian2025uavs} exploiting the powerful visual reasoning capabilities of VLM to address the challenges posed by UAV navigation scenes. Despite the progress, the generalizability and performance of these methods still require improvement.

\paragraph{LLMs for Robot Planning and Interaction.}

Most recently, LLMs have demonstrated impressive capabilities in understanding and reasoning. To leverage these capabilities, several promising methods have been proposed for applying LLMs in robotic systems. A few methods involve using LLM-generated rewards optimized in simulation to improve control~\cite{huang2023voxposer, yu2023language}. Others utilize LLM-selected subgoals as an abstraction to enhance policies for navigation~\cite{dorbala2022clip, chen2023open} and manipulation~\cite{cui2022can, li2024embodied}. Additionally, research has explored the use of LLMs to generate executable code for control and perception primitives~\cite{singh2023progprompt, liang2023code, li2025visual}. Despite their potential, LLMs are still prone to confidently hallucinating outputs, such as referring to objects not observed in the scene~\cite{zeng2022socratic}. 
In order to alleviate the hallucinating phenomenon, we propose a matrix-based representation containing topological, semantic, and metric information for better prompting the LLMs.

\section{Method}
\label{sec:approach}



\begin{figure}[!t]
    \vspace{-5pt}
    \centering
    \includegraphics[width=1.0\linewidth]{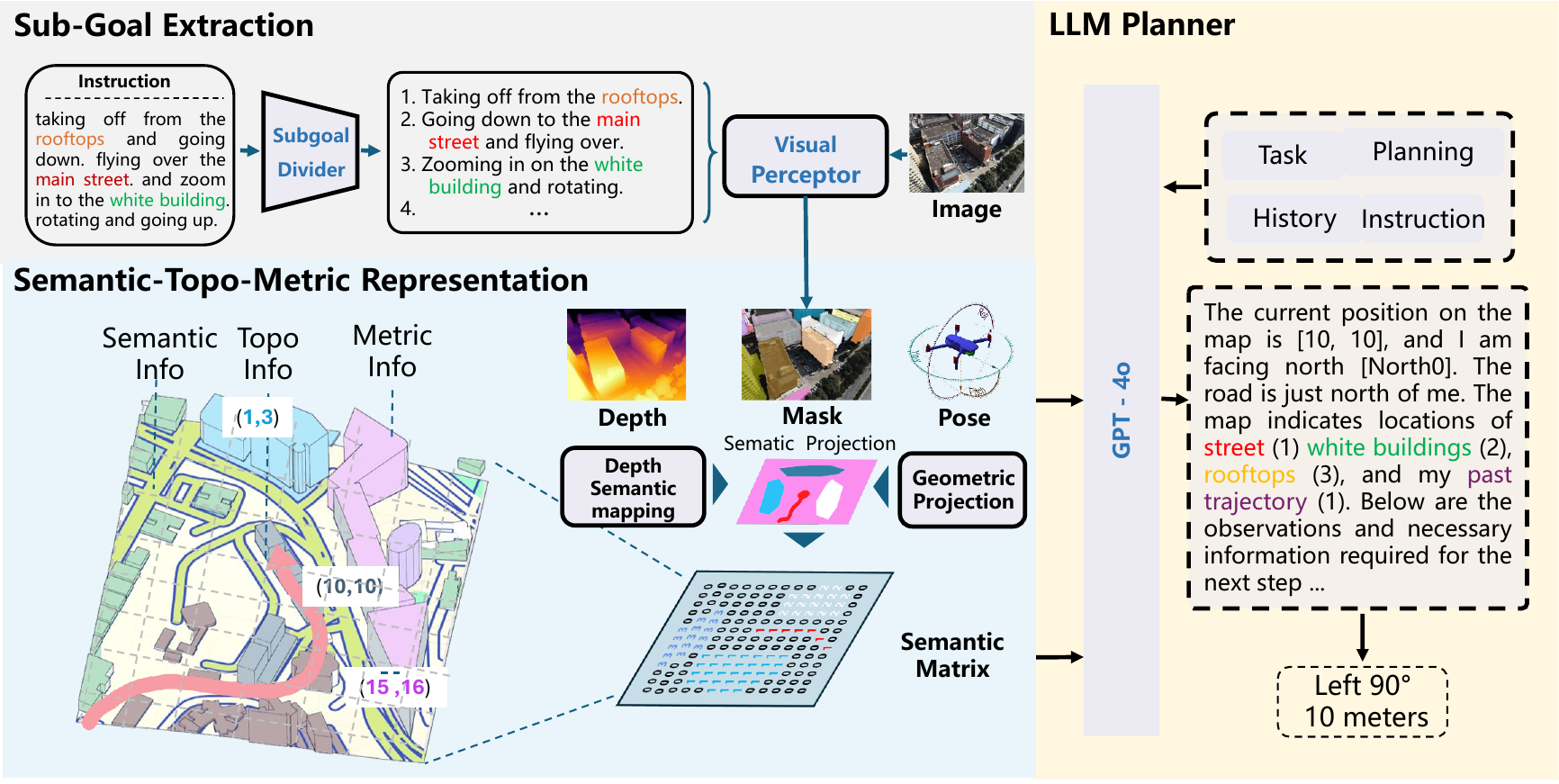}
    \caption{Our method consists of three modules, \emph{i.e.}, Sub-Goal Extraction, Semantic-Topo-Metric Representation, and LLM planner. They are utilized to generate sub-goal instructions, spatial information representations, and UAV navigation actions, respectively. }
    \label{fig:overview}
\end{figure}

%
In this paper, we propose a zero-shot framework that leverages Large Language Models (LLMs) for action prediction in aerial VLN tasks. As shown in \cref{fig:overview}, our framework consists of three modules. The sub-goal extraction module decomposes language instructions into several sub-goals, facilitating step-by-step reasoning and navigation. The Semantic-Topo-Metric Representation (STMR) module represents the outdoor environment as a matrix containing semantic, topological, and metric information, enhancing the spatial reasoning ability of LLMs. Finally, an LLM planner is designed, which takes sub-goal instructions, STMR, together with historical information and task description as input, and outputs its thoughts and predicted actions.

\subsection{Problem Formulation}

The aerial VLN task is formulated as a free-form language instruction guided navigation. At the beginning of each episode, the initial pose of a UAV is denoted as $P = [x, y, z, \phi, \theta, \psi]$, where $(x, y, z)$ is the UAV's position and $(\phi, \theta, \psi)$ represents pitch, roll, and yaw of the UAV's orientation. A natural language instruction $L$ is provided to specify the path that a UAV should follow. 
%
To achieve the navigation goal, the UAV considers both the instruction and visual perceptions, and predicts an action from the action space (\emph{i.e.}, right, left, up, down, forward, backward) with corresponding value (\emph{e.g.}, 1-10m)  at each time step $t$.
Navigation ends when the UAV predicts a \textit{`Stop'} action or reaches a pre-defined maximum action number. Following~\cite{liu_2023_AerialVLN}, the navigation is considered successful if the UAV halts within 20 meters of the target location.


\subsection{Semantic-Topo-Metric Representation (STMR)}
\label{sec:hybrid}

%

Previous LLM-based VLN methods use natural language to describe current observations, or a topological graph to model the spatial information of the environment. However, in open scenarios, simple directional words such as ``next to" or ``aside" are insufficient for describing complex spatial relationships, often leading to ambiguity in LLMs.
To address this challenge, we introduce the STMR to enhance the spatial-aware reasoning capability of LLMs. Specifically, STMR incrementally takes an RGB image $I_t^R$ and a depth map $I_t^D$ as input from each step, and generates a dynamically updated matrix representation with semantic, topological, and metric information as its output. The details of STMR are presented as follows.

\subsubsection{2D Visual Perception.}
Impressed by the powerful open-vocabulary detection capabilities of Grounding DINO $\operatorname{GD}(\cdot)$, as well as the captioning and segmentation capabilities of Tokenize Anything model $\operatorname{TA}(\cdot)$, we integrate these two models as our 2D visual preceptor, as illustrated in \cref{fig:2DVP}. Given a single RGB image $I_t^R$ and an instruction $L$ as input, we first obtain detailed landmark categories $C = \{c_1, c_2, ..., c_n\}$ using a Landmark Extractor $\operatorname{LE}(\cdot)$ driven by an LLM, and then identify the corresponding bounding box for each category through $\operatorname{GD}(\cdot)$.
Next, we employ $\operatorname{TA}(\cdot)$ to take each bounding box as a prompt and output a set of 2D semantic masks $m^{(t)}$ and captions $h^{(t)}$ for the current RGB image $I_t^R$. The entire process can be described as:
\begin{figure}[!t]
\vspace{-5pt}
    \centering
    \includegraphics[width=1.0\linewidth]{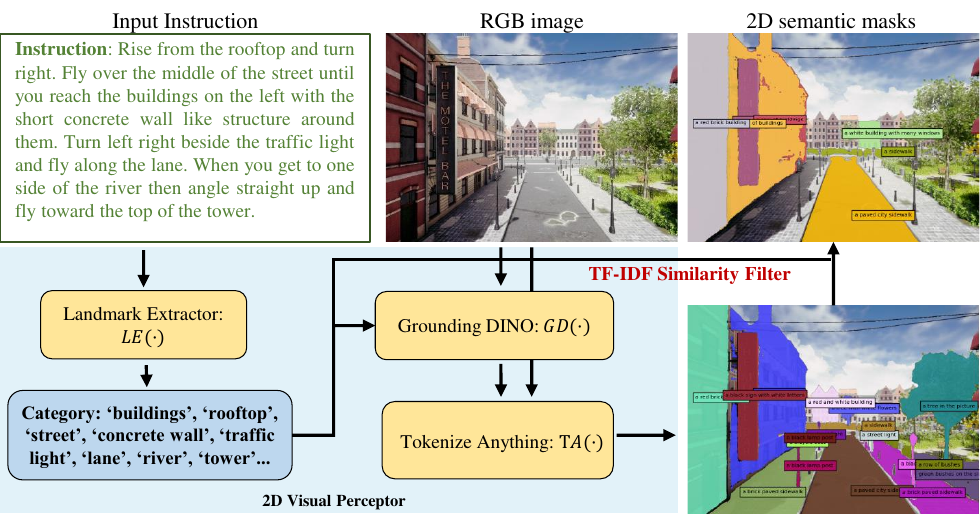}
    \caption{2D Visual Perceptor for the UAV.}
    \label{fig:2DVP}
    \vspace{-10pt}
\end{figure}

\begin{equation}
    \label{cap}
    \{m^{(t)}, \  h^{(t)}\}=\operatorname{TA}\left( {I_t^R}, \  
    \operatorname{GD}\left( {I_t^R}, \ \operatorname{LE}({L}) \right) \right). \\
\end{equation}

To improve the robustness of the semantic masks and reduce the misleading of LLM reasoning caused by numerous open-vocabulary categories, we propose a text-matching method to mitigate over-classification. 
As shown in \cref{fig:2DVP}, our method involves vectorizing the landmarks extracted from the instruction and the captions $h^{(t)}$ generated in each $I_t^R$. 
Then, we calculate the cosine similarity between these vectorized landmarks and captions using TF-IDF~\cite{qaiser2018text}. If the similarity score exceeds the threshold $\tau$ (0.8), the landmark is classified as visible in the current view, and other irrelevant masks will be ignored. 
By implementing this strategy, we effectively simplify semantic masks and ensure the LLM's reasoning focuses on relevant categories.




\subsubsection{Sub-goal-driven Top-down Map.}
 Considering that a top-down view better expresses spatial relations, we use the depth image to map semantic masks to the 3D space, and then project them into a top-down map. Note that the top-down map grows and updates during the UAV flight to support the navigation process. Specifically, by applying the above 2D visual perception process to the continuously observed RGB images, we obtain segmented images with identified objects and regions. Subsequently, the corresponding depth images are projected to a gradually growing 3D point cloud, where each pixel is mapped to a 3D point $(X, Y, Z)$ based on its depth value and camera parameters:

\begin{gather}
X=\frac{(u - c_x) I_t^D(u, v)}{f_x},
Y=\frac{(v - c_y) I_t^D(u, v)}{f_y}, \\
Z= I_t^D(u, v),
 \label{icp}
\end{gather}
where $(u, v)$ are the pixel coordinates, $(c_x, c_y)$ are the camera's principal point coordinates, and $(f_x, f_y)$ are the focal lengths.
The semantic labels from the segmented image are mapped to the corresponding 3D points, resulting in a point cloud with semantic information $(X, Y, Z, C_i)$, where $C_i$ is the semantic category.
Then the 3D point cloud is partitioned into discrete voxels, where each voxel aggregates its point clouds as one semantic category using max pooling.
For a specific coordinate $(x, y)$, sometimes different categories of objects will appear at different heights. For example, there is `vegetation' or `equipment' on the roof of a `building'. Given that UAVs usually fly above the landmarks, the semantic label of the top one in a column of voxels will be projected into the top-down map:
\begin{equation}
\text{TopDownMap}(x, y) = \text{Voxel}(x, y, z_{top}),
\label{eq:proj}
\end{equation}
where $z_{top}$ means the highest $z$ coordinate at location $(x,y)$ and $\text{Voxel}(x, y, z_{top})$ denotes the corresponding semantic label. Consequently, we get a top-down map with semantic information. Notably, since the LLM prioritizes the landmarks in the current sub-goal, if a category contained in the sub-goal appears in the voxel at any $z$ coordinate, this category will be projected into the top-down map first. Thus we can modify \cref{eq:proj} as:

\begin{align}
\text{TopDownMap}(x, y) = 
\begin{cases} 
C_i, & C_i \text{ in sub-goal} \\
\text{Voxel}(x, y, z_{top}), & \text{otherwise}.
\end{cases}
\end{align}

\subsubsection{Matrix Representation.} 
We found that directly inputting the image-format top-down map to a vision-language model (VLM) often yields poor reasoning results. Alternatively, we process the visual map into an LLM-friendly text-based matrix representation, which effectively guides the navigation.
In order to make LLMs aware of metric information, we define a 100m $\times$ 100m local map centered on the UAV's current position from the entire top-down map, and divide it into $20\times20$ coordinate grids with each grid covering a 5m $\times$ 5m area. The grid size is selected according to the size of the most common small landmarks (\emph{i.e.}, cars) to ensure them occupy a single grid, while larger landmarks can be represented by multiple grids. Subsequently, a text-based matrix is generated by applying semantic max pooling in each grid, where the most frequent category is selected as the semantic label. Each category, such as building or car, is identified by different number labels. Furthermore, the grid size $r$ (5m) is also input to the LLM as the metric information. Notably, the local map slides dynamically during the UAV flight, enhancing the LLM to understand the surrounding environment and reason about positional relationships among landmarks.


\subsection{LLM-based Navigator}
\label{sec:prompting}
To improve the robustness of the LLM-based navigator, the prompt mainly consists of two components, \emph{i.e.}, task definition and STMR-based planner. Firstly, the task definition includes a description of the navigation task, as well as the input and output format. The input format specifies the size of the matrix representation and defines the mapping relationship between the semantic labels and matrix elements. The output format outlines the action space, \emph{i.e.}, \textit{(right, left, lift, down, straight, back)}, along with the maximum moving distance and turning range. Secondly, the STMR-based planner includes a text-based matrix representing contextual observation, sub-goal-based next-step planning, and history actions. The LLMs are required to leverage Chain-of-Thought \cite{wei2022chain} reasoning after capturing the observation, thinking step-by-step in the order of observation-thought-planning-history to predict the next actions. During the navigation process, the status of each sub-goal is updated, consisting of three states, \emph{i.e.}, \textit{todo}, \textit{in process}, and \textit{completed}, to further alleviate the issue of LLM hallucinations. The text-based matrix is progressively updated as the UAV navigates through the environment. The prompt examples can be found in the supplementary material.

\section{Experiment}
\label{sec:experiment}

\subsection{Dataset and Implementation Details} 
\subsubsection{Dataset.}
We conducted experiments using two test sets, \emph{i.e.}, the simple navigation test set collected by us and the complex navigation test set from the AerialVLN-S benchmark~\cite{liu_2023_AerialVLN}. Both test sets are derived from Microsoft AirSim plugins and Unreal Engine 4 scenes. Our simple test set consists of 1,000 trajectories that cover a range of common navigation scenarios, \emph{e.g.}, public transportation and urban environments. Each trajectory spans up to 100 meters and contains 1 or 2 landmarks as navigation references. This test set will also be released with the code. AerialVLN-S replicates real-world urban environments with over 870 object categories across diverse scenarios, including downtown cities, factories, parks, and villages. The flight trajectories are collected by AOPA-certified UAV pilots and curated by experts to ensure realism and navigational fidelity. For the AerialVLN-S benchmark, the trajectory is much longer (326.9 meters on average) with more landmarks and more complex instructions (83 words per instruction on average), causing an extremely challenging task. Besides, to verify the robustness and generalizability of the proposed method, we conduct real-world experiments in 10 outdoor scenes.

\subsubsection{Evaluation Metrics.}
To validate the effectiveness of our method, we utilize a comprehensive set of evaluation metrics following \cite{liu_2023_AerialVLN}. We focus on several key aspects, \emph{i.e.,} Navigation Error (NE), quantifying the distance between the UAV's stopping point and the actual destination; Success Rate (SR), measuring the proportion of navigations that successfully reach the destination within a 20-meter threshold; Oracle Success Rate (OSR), an idealized measure considering any point on the predicted trajectory that comes within 20 meters of the destination as a success; Normalised Dynamic Time Warping (SDTW), taking into account both the navigation success rate and the similarity between the predicted trajectory and ground truth.

\subsubsection{Implementation Details.}
Our framework is implemented in both a simulator and a real environment. In the simulation environments of AirSim and UE4, we evaluated the methods using a PC equipped with an Intel i9 12th-generation CPU and Nvidia RTX 4090 GPUs. For the real world environment, we test on a Q250 airframe, carrying an Intel RealSense D435i depth camera, and an NVIDIA Jetson Xavier NX running Ubuntu 18.04 as the onboard computer. This setup supports the construction of top-down maps and flight control. The deployment of large-scale models and communication with the onboard computer are managed by a mobile ground station equipped with high-performance computing capabilities (matching the simulation setup). For LLM reasoning, we utilize the online API of GPT-4o, employing the default parameters.

\begin{figure}[!t]
    \centering
    \includegraphics[width=\linewidth]{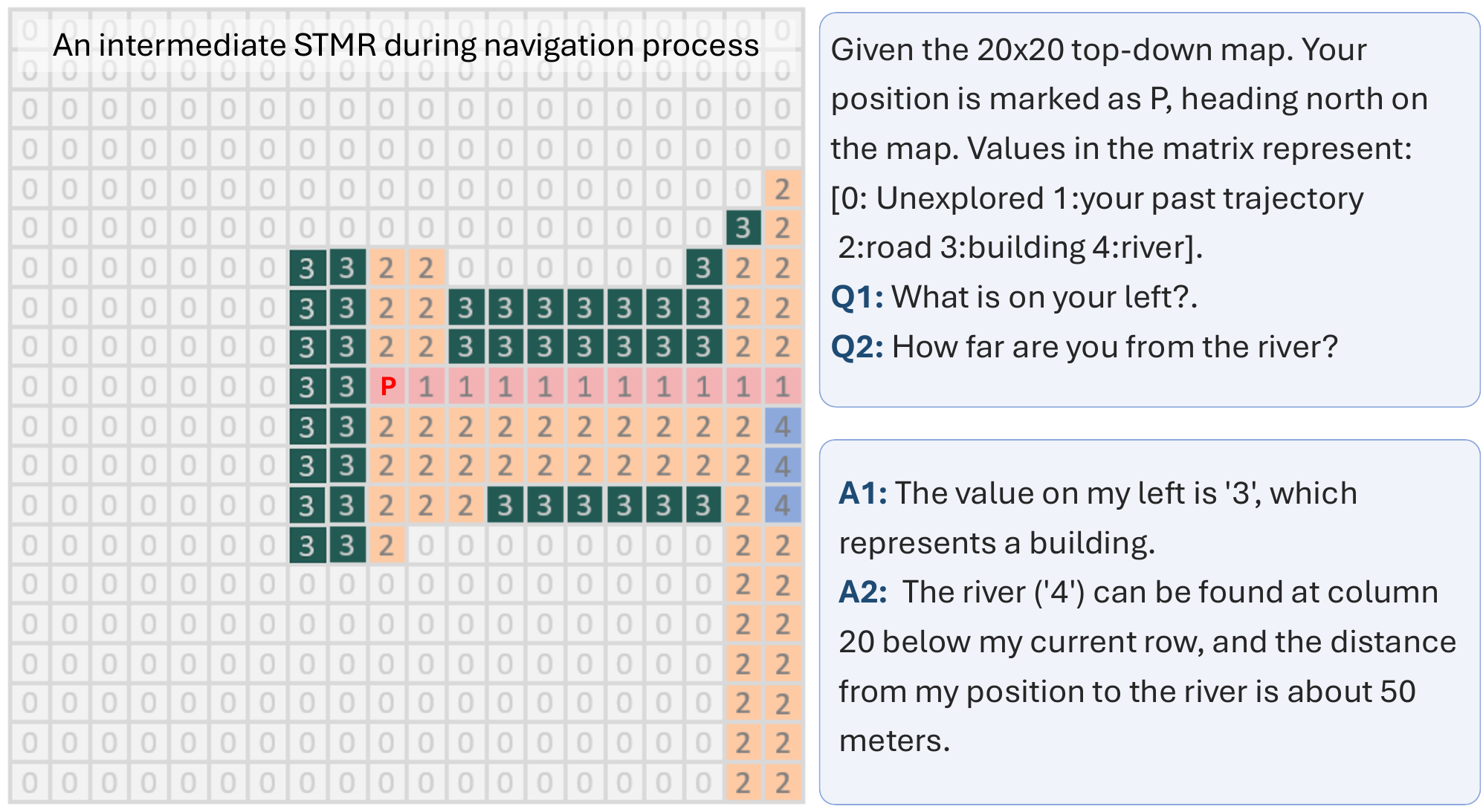}
    \caption{Demonstration of the STMR in spatial reasoning.}
    \label{fig:stmr_spatial_reasoning}
    \vspace{-2mm}
\end{figure}

\begin{table}[t]
\centering 
\resizebox{.95\columnwidth}{!}{
\begin{tabular}{c|cccc}
\hline
\multicolumn{1}{c|}{\textbf{model}} & NE/m $\downarrow$ & SR/\% $\uparrow$ & OSR/\% $\uparrow$ & SDTW/\% $\uparrow$  \\ \hline

 Random                                           & 266.7             & 2.0              & 5.9      &  0.2          \\
 Action Sampling                                  & 259.9             & 5.1             & 8.0      &  0.3      \\   
 Seq2Seq~\cite{anderson2018vision}                & 130.1             & 25.5              & 57.1     &  10.6      \\
 CMA~\cite{CMA}                                   & 113.6             & 29.8              & 73.2     &  10.5           \\
 AerialVLN~\cite{liu_2023_AerialVLN}              & 116.4                & 37.7              & 71.7     &  12.7     \\
 Navid~\cite{navid}                              & 99.5               & 44.1                & 79.9        &  16.6    \\ 
OpenUAV~\cite{openuav}                         & 108.8    & 49.9   & 68.0    &  14.9  \\
 \textbf{Ours (training-free)}                                    & \textbf{53.9}              &  \textbf{76.7}   & \textbf{90.3}     &  \textbf{18.0}    \\\hline
\end{tabular}
}
\caption{Overall performance comparisons on simple tasks.}
\label{tab:baseline_self}
\end{table}


\begin{table}[t]
\centering 
\resizebox{.95\columnwidth}{!}{
\begin{tabular}{c|cccc}
\hline
 \multicolumn{1}{c|}{}      & \multicolumn{4}{c}{\textbf{AerialVLN-S}}           \\
\multicolumn{1}{c|}{\multirow{-2}{*}{\textbf{model}}} & NE/m $\downarrow$ & SR/\% $\uparrow$ & OSR/\% $\uparrow$ & SDTW/\% $\uparrow$  \\ \hline

 Random                                           & 109.6             & 0.0              & 0.0      &  0.0          \\
 Action Sampling                                  & 213.8             & 0.9              & 5.7      &  0.3      \\
 LingUNet~\cite{misra2018mapping}                 & 383.8             & 0.6              & 6.9      &  0.2      \\
 Seq2Seq~\cite{anderson2018vision}                & 146.0             & 4.8              & 19.8     &  1.6      \\
 CMA~\cite{CMA}                                   & 121.0             & 3.0              & 23.2     &  0.6           \\
 AerialVLN~\cite{liu_2023_AerialVLN}              & \textbf{90.2}     & 7.2              & 15.7     &  2.4     \\
 Navid~\cite{navid}                               & 105.1               & 6.8                & 15.5        &  1.1    \\ 
OpenUAV~\cite{openuav}                         & 102.8             & 6.3             & 17.6      &  2.0                          \\
 \textbf{Ours (training-free)}                                    & 96.3              &  \textbf{12.6}   & \textbf{31.6}     &  \textbf{2.7}    \\\hline
\end{tabular}
}
\caption{Comparison on the validation seen set of AerialVLN-S.
}
\label{tab:baseline_seen}
\end{table}

\subsection{Experimental Results}
\subsubsection{Baseline Models}
\begin{itemize}[leftmargin=*]
\item[$\bullet$] \textbf{Rule-based Methods.} Rule-based methods include Random and Action Sampling. In the Random method, the agent randomly chooses actions at each step and continues until either the `stop' action is chosen or the maximum number of steps is reached. The Action Sampling method, on the other hand, first analyzes the statistical properties of the dataset and then samples actions based on the action distribution.

\item[$\bullet$] \textbf{Learning-based Methods.} Learning-based methods include mainstream aerial navigation approaches like LingUNet, Seq2Seq and CMA~\cite{liu_2023_AerialVLN}, as well as the state-of-the-art method Navid~\cite{navid} and OpenUAV~\cite{openuav}. Different from other baselines, Navid is a method for the indoor VLN task, so we retrain it on the original AerialVLN dataset.

\item[$\bullet$] \textbf{LLM-based Methods.} For LLM-based methods, we compare two works for the indoor VLN task, \emph{i.e.}, MapGPT \cite{chen2024mapgpt} and NavGPT \cite{zhou2024navgpt}. To ensure fairness, all these methods are evaluated using the GPT-4o with the same setting.
\end{itemize}

\subsubsection{Quantitative Results in Simulator.} 
We first analyze performance on the simple navigation tasks. The scenes mainly tests the agent's ability to locate landmarks and predict actions correctly. Even a short instruction like ``fly along the road and stop near the intersection" requires the UAV to accurately understand the scene, recognize the destination, and stop in a proper location. As shown in~\cref{tab:baseline_self}, our method outperforms other works by a large margin in terms of SR (+26.8\%) and OSR (+22.3\%). In addition, our method exhibits a significantly lower navigation error (NE) compared to the other methods, which is reduced by 45.6 meters on average compared with the second-place method, Navid. These indicate that the UAV has a better understanding of its own spatial position and semantic constraints under the guidance of our STMR.
In~\cref{tab:baseline_seen}, the proposed method consistently achieves superior performance on the AerialVLN-S dataset. As the task complexity increases significantly with much longer trajectories and instructions, the success rates of all methods experience a notable decline.  The results in~\cref{tab:baseline_seen} indicate that Aerial VLN remains a challenging task having not been fully studied, and there is still room for improvement in overall performance. We believe that our work can bring valuable insights into this field.
%

\subsubsection{Quantitative Results in Real Environment.}   
To test the performance of our solution in the real world, we collected 10 outdoor scenes, including street scenes and forests, with ground-truth lengths ranging from 50m to 200m. Then, We evaluate our method using a real UAV, leveraging cloud-hosted LLMs to navigate through these challenging environments. In this experiment, we employ and compare not only pretrained VLN models but also other LLM-based VLN methods, \emph{i.e.,} MapGPT~\cite{chen2024mapgpt} and NavGPT~\cite{zhou2024navgpt}. Notably, MapGPT and NavGPT rely on predefined topological maps to predict the next navigation waypoint. In this experiment, we manually define candidate waypoints for both methods.

\begin{figure*}[!t]
    \centering
    \includegraphics[width=0.9\linewidth]{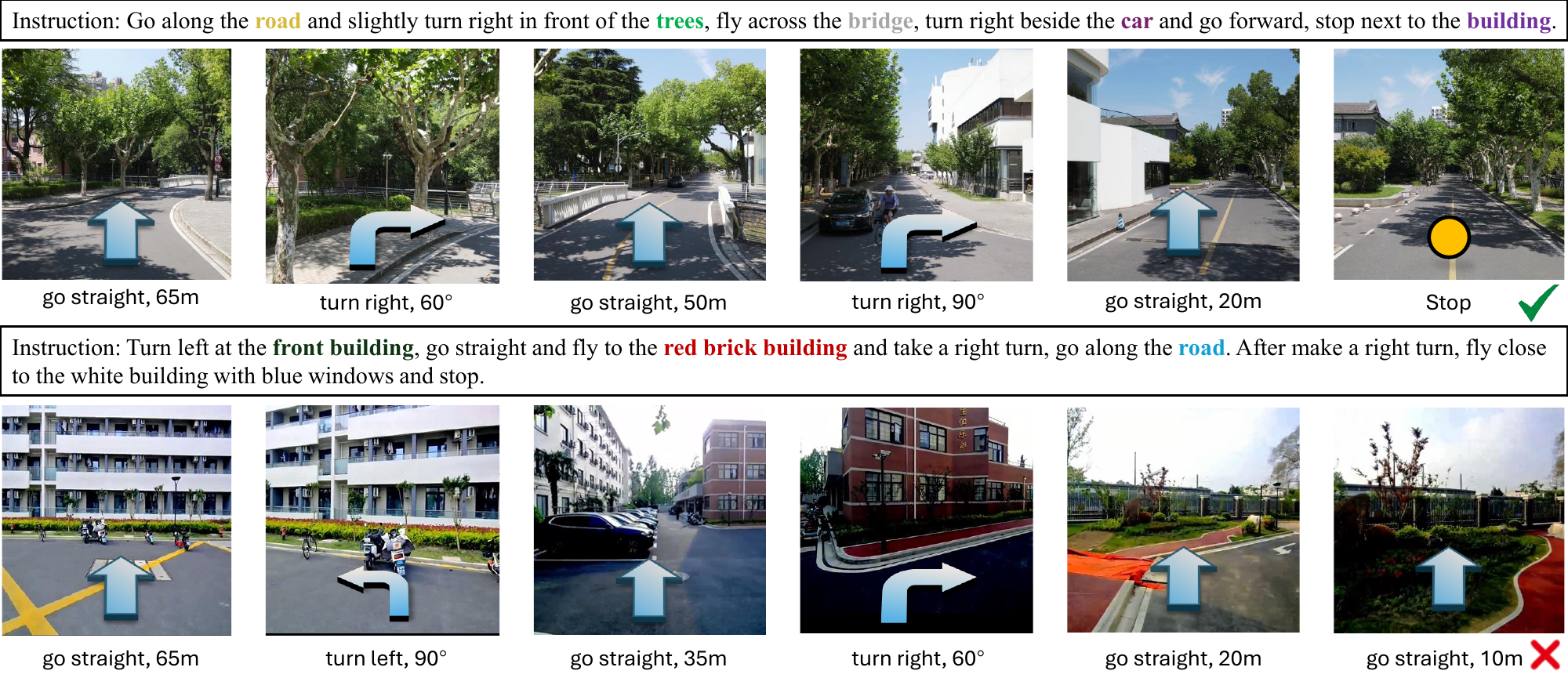}
    \caption{Visualization of successful and failed examples in the real environments. Short-range movements of the same action are merged into a single long-range movement, as shown in the image sequence. The second row illustrates a failed case, the UAV mistakenly executed the ``go straight" action because the visual perceptor failed to recognize the road in the fifth image.}
    \vspace{-2mm}
    \label{fig:demo}
\end{figure*}

As shown in the demonstration snapshot in~\cref{fig:demo}, our method effectively aligns visual and textual landmarks, understands commands, and successfully guides the UAV to its destination. ~\cref{tab:real} also shows that the proposed method achieves a much better performance than both pretrained VLN methods and training-free LLM-based methods. However, we can see that LLM-based methods, including ours, need more computation time. This is mainly because of the latency of the LLM API. We believe that with the development of LLMs and hardware computing power, the latency will gradually decrease.

\begin{figure}[!t]
    \centering
    \includegraphics[width=\linewidth]{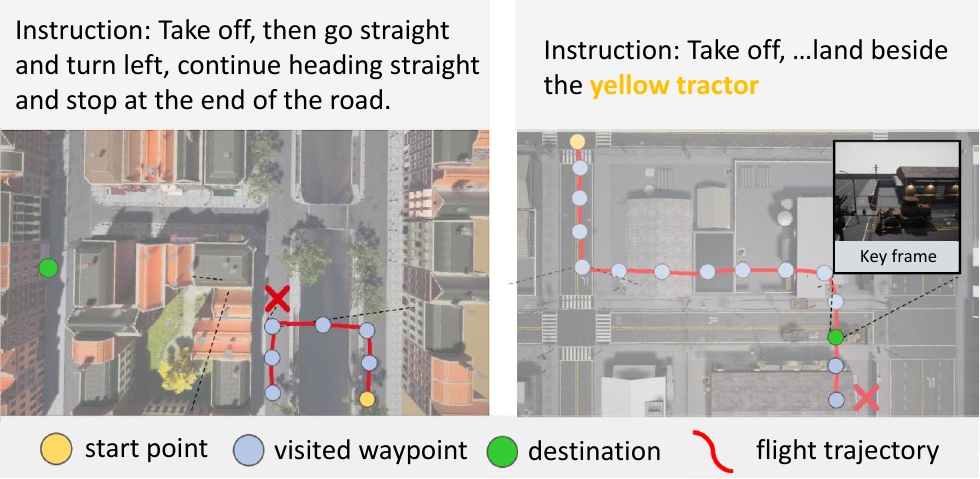}
    \caption{Failure cases of our method.}
    \label{fig:stmr_failure_case}
\end{figure}

\begin{table}[t]
    \begin{center}
    \resizebox{.95\columnwidth}{!}{\begin{tabular}{c|ccc} 
\hline  \multicolumn{1}{c|}{\textbf{model}} 
 & SR/\% $\uparrow$ & OSR/\% $\uparrow$ & Time (s) \\ 
    \hline
    AerialVLN~\cite{liu_2023_AerialVLN} & 0  & 10  &  82  \\     
    Navid~\cite{navid} & 10 & 10 &  75  
    \\ 
    MapGPT~\cite{chen2024mapgpt} & 20 & 20 &  176  \\ 
    NavGPT~\cite{zhou2024navgpt} & 10 & 20 &  150  \\  
    Ours    & \textbf{40} & \textbf{70}   &  117 
    \\
    \hline
    \end{tabular}}
    \caption{Quantitative results in real environment. The last column shows the average runtime of aerial VLN procedures.}
    \label{tab:real}
    \end{center}
    \vspace{-2mm}
\end{table}

\subsubsection{Case Analysis.} 

\cref{fig:stmr_spatial_reasoning} demonstrates the effectiveness of our STMR in helping spatial reasoning, where the left part shows an intermediate STMR during the navigation process, and the right part shows simplified prompt, questions, and reasoning answers. As shown in \cref{fig:stmr_spatial_reasoning}, an LLM can understand and describe the surrounding environment through the STMR. It can even make accurate judgments about objects or areas at a distance. This ability is particularly valuable for aerial VLN in vast outdoor environments.

\cref{fig:stmr_failure_case}  illustrates two of the most common failure cases caused by incorrect planning or execution. The first typical failure arises from the misunderstanding of ambiguous instructions. there are many continuous commands without any landmarks, such as \emph{turn left, then move right, then go straight}, which lack landmarks for the LLM to reference. This ambiguity often leads to the repeated execution of an action.
The second typical error is caused by the inaccuracy of visual perception. Although modern perception models show strong capability, they still require improvement in identifying objects from different views. As a result, the key landmark may not be mapped to the STMR, causing the proper action not to be completed as illustrated in the right part of \cref{fig:stmr_failure_case} and the second row of~\cref{fig:demo}, showcasing the challenges posed by complex environments.

\begin{table}[!ht]
	\centering
\renewcommand\arraystretch{1.3}
	\setlength{\tabcolsep}{3mm}

        \resizebox{.95\columnwidth}{!}{
	\begin{tabular}{c|cccc}
        \hline
		\multirow{2}*{Method}  & \multicolumn{4}{c}{\textbf{Validation Unseen}}  \\  
                        & NE/m $\downarrow$     & SR/\% $\uparrow$    & OSR/\% $\uparrow$  & SDTW/\% $\uparrow$   \\ \hline
     Topo                  & 203.3        &  4.9         &  12.8   &  1.8  \\
     Metric                    & 165.0         & 6.1         & 13.7          &  2.2   \\
     Ours                  & \textbf{88.7}    & \textbf{15.0}     & \textbf{28.0}     &  \textbf{3.6}    \\ \hline
    \end{tabular}
    }
\caption{Ablation study on different spatial representations.}
\label{tab:different_spatial_info}
\vspace{-3mm}
\end{table}

\subsection{Ablation Studies}
We conduct comprehensive ablation studies to assess the core components of the proposed method. We randomly sample 100 samples from the unseen validation dataset of AerialVLN-S and perform all the ablation experiments. Corresponding results are depicted in~\cref{tab:different_spatial_info}, ~\cref{tab:visual_vs_stmr}, ~\cref{fig:grid_size}, and~\cref{tab:state_update}.

\subsubsection{Different Types of Spatial Information.}
To demonstrate the spatial representation ability of STMR, we further compare it with other prompting formats as an LLM's spatial information prompt. In Topo format (row 1 of~\cref{tab:different_spatial_info}), we maintain a linguistically formed map that captures the topological relationships between different nodes. Each node records textual descriptions of visual observations, and the connectivity between nodes is described using textual prompts. For example, \emph{Place 1 is connected with Places 2, 4, 0, 3}. In the Metric format (row 2 of~\cref{tab:different_spatial_info}), we arrange the visual observations from 8 different directions in a clockwise order relative to the agent's current orientation and concatenate them into a single prompt. Each visual observation records the direction and distance of the landmark, for example, \emph{a white building in the left front 10 meters away}. It is shown that row 2 improves SR by 1.2\% (absolute) over row 1 but still demonstrates poorer navigation performance. This is because, for certain landmarks like roads and rivers, UAVs still cannot obtain region-level spatial information based solely on point-distance descriptions. In contrast, our STMR combining the semantic, topological, and metric information, significantly enhances the UAV’s exploration capability and improves OSR accuracy by over 10\%.

\subsubsection{Direct Visual Input vs STMR.}
Due to the powerful multimodal representation capabilities of GPT-4V and GPT-4o, they can directly reason with both visual and textual prompts. However, it is challenging for them to make correct VLN decisions from sequential images. This may be attributed to the inability of VLMs in spatial-aware reasoning. As shown in~\cref{tab:visual_vs_stmr}, the proposed STMR representation with LLMs significantly outperforms the direct input of RGB images into VLMs. It demonstrates that LLMs can get better spatial reasoning abilities when provided with proper visual encoding.

\subsubsection{Grid Size of STMR.}
\begin{figure}[!t]
    \centering
    \includegraphics[width=0.5\linewidth]{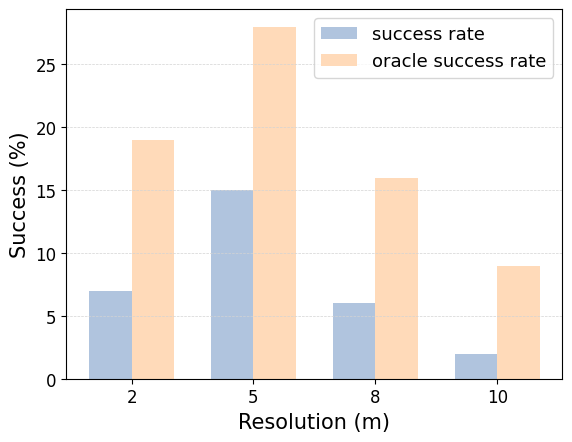}
    \caption{Ablation study on different STMR grid sizes.}
    \label{fig:grid_size}
\end{figure}

In \cref{fig:grid_size}, we demonstrate the necessity of maintaining a trade-off between grid resolution and the total number of grid cells in STMR representation. Set the resolution to 5m may not allow for as precise spatial inference as 2m, but it is enough for outdoor spatial reasoning. 

\subsubsection{State Update of Sub-goal.}
To ensure that each sub-goal can be executed despite environmental changes, the LLM does not change the original path plan in each iteration. Instead, it updates the status of the current sub-goal, including \textit{todo}, \textit{in process}, and \textit{completed}. We compare this strategy with one that updates the entire multi-step path planning (row 1 of \cref{tab:state_update}) based on the observation in each iteration.
This strategy does not notably enhance the OSR performance, indicating that the proposed STMR already furnishes enough capacity for navigating the environment comprehensively. Nonetheless, it does positively impact the decision-making process, leading to an increase in the success rate (SR) from 9.0\% to 15.0\%. 

\begin{table}[!ht]
    \begin{center}
    \resizebox{.90\columnwidth}{!}{\begin{tabularx}{1.25\linewidth}{c|p{0.5cm}p{0.8cm}|cccc}
    \hline
    \diagbox[width=1.3cm, height=0.9cm]{\footnotesize VLM}{\footnotesize Input} & RGB & STMR & NE/m $\downarrow$      & SR/\% $\uparrow$    & OSR/\% $\uparrow$  & SDTW/\% $\uparrow$  \\ \hline
    \multirow{2}{*}{GPT-4V}          & \textbf{$\checkmark$}  &            & 350.0          & 1.5          & 9.7    & 0.4    \\
      &     & \textbf{$\checkmark$}        & 112.5          & 10.2          & 22.4    & 2.1    \\ \hline
    \multirow{2}{*}{GPT-4o} & \textbf{$\checkmark$}  &            & 412.0          & 1.1          & 10.0  & 0.2    \\
     &     & \textbf{$\checkmark$} & \textbf{88.7}    & \textbf{15.0}     & \textbf{28.0}     &  \textbf{3.6}    \\ \hline
    \end{tabularx}}
    \caption{Ablation study on the different visual prompts for VLMs.}
    \label{tab:visual_vs_stmr}
    \end{center}
    \vspace{-4mm}
    
\end{table}

\begin{table}[t]
    \begin{center}
    \resizebox{.85\columnwidth}{!}{\begin{tabular}{c|cccc} 
\hline
Module & NE/m $\downarrow$ & SR/\% $\uparrow$ & OSR/\% $\uparrow$ & SDTW/\% $\uparrow$ \\ 
    \hline
    w/o state & 289.7 & 9.0 & 22.1 & 1.7 \\ 
    w state & \textbf{88.7}    & \textbf{15.0}     & \textbf{28.0}     &  \textbf{3.6}    \\  
    \hline
    \end{tabular}}
    \caption{Ablation study on the state update strategies.}
    \label{tab:state_update}
    \end{center}
    \vspace{-5mm}
    
\end{table}


\section{Conclusions}
\label{sec:conclusion}
This paper addresses the challenging aerial VLN task by proposing an LLM-based training-free framework. To enhance the spatial reasoning ability of LLMs, we design Semantic-Topo-Metric Representation (STMR). STMR first integrates instruction-related landmarks and their locations into a top-down map, and subsequently transforms this map into a matrix representation containing semantic, topological, and distance metric information. Taking the proposed STMR as a part of the LLM prompts, we significantly improve the UAV's navigation capabilities. Our framework achieves state-of-the-art results on both simple navigation tasks and complex navigation tasks, demonstrating its effectiveness and robustness.
\section{Limitations and Future Work}
\label{sec:limitation}
While the proposed STMR significantly improves LLM's spatial understanding ability, it still struggles to handle very precise distances in 3D space. Additionally, pretrained perception models sometimes introduce errors under different observation angles and distances, causing difficulties for subsequent navigation. Future avenues of improvement include using the primary viewpoint to assist spatial perception, or designing more robust perception strategies.

\bibliography{bibliography}
\section{Reproducibility Checklist}

\begin{enumerate}
    \item This paper:
    \begin{itemize}
        \item Includes a conceptual outline and/or pseudocode description of AI methods introduced (\textbf{\underline{yes}}/partial/no/NA)
        \item Clearly delineates statements that are opinions, hypotheses, and speculation from objective facts and results (\highlight{yes}/no)
        \item Provides well-marked pedagogical references for less-familiar readers to gain background necessary to replicate the paper (\highlight{yes}/no)
    \end{itemize}

    \item Does this paper make theoretical contributions? (\highlight{yes}/no) \\
    If yes, please complete the list below:
    \begin{itemize}
        \item All assumptions and restrictions are stated clearly and formally. (\highlight{yes}/partial/no)
        \item All novel claims are stated formally (e.g., in theorem statements). (\highlight{yes}/partial/no)
        \item Proofs of all novel claims are included. (\highlight{yes}/partial/no)
        \item Proof sketches or intuitions are given for complex and/or novel results. (\highlight{yes}/partial/no)
        \item Appropriate citations to theoretical tools used are given. (\highlight{yes}/partial/no)
        \item All theoretical claims are demonstrated empirically to hold. (\highlight{yes}/partial/no/NA)
        \item All experimental code used to eliminate or disprove claims is included. (\highlight{yes}/no/NA)
    \end{itemize}

    \item Does this paper rely on one or more datasets? (\highlight{yes}/no) \\
    If yes, please complete the list below:
    \begin{itemize}
        \item A motivation is given for why the experiments are conducted on the selected datasets. (\highlight{yes}/partial/no/NA)
        \item All novel datasets introduced in this paper are included in a data appendix. (\highlight{yes}/partial/no/NA)
        \item All novel datasets introduced in this paper will be made publicly available upon publication with a license allowing free research use. (yes/partial/no/\highlight{NA})
        \item All datasets drawn from the existing literature are accompanied by appropriate citations. (\highlight{yes}/no/NA)
        \item All datasets drawn from the existing literature are publicly available. (\highlight{yes}/partial/no/NA)
        \item Datasets that are not publicly available are described in detail, with justification. (yes/partial/no/\highlight{NA})
    \end{itemize}

    \item Does this paper include computational experiments? (yes/no) \\
    If yes, please complete the list below:
    \begin{itemize}
        \item Number/range of values tried per (hyper-)parameter and selection criteria are reported. (yes/partial/no/NA)
        \item Code for data preprocessing is included in the appendix. (yes/partial/no)
        \item Source code for conducting and analyzing experiments is included. (yes/partial/no)
        \item Code will be released publicly upon publication with a permissive license. (yes/partial/no)
        \item Code includes comments with implementation details and paper references. (yes/partial/no)
        \item Seed setting methods for stochastic algorithms are described. (yes/partial/no/NA)
        \item Computing infrastructure (hardware/software specs) is reported. (yes/partial/no)
        \item Evaluation metrics are formally described with motivations. (yes/partial/no)
        \item Number of runs per result is specified. (yes/no)
        \item Performance analysis includes variation, confidence, or distributions. (yes/no)
        \item Significance of performance differences is assessed with statistical tests. (yes/partial/no)
        \item Final (hyper-)parameter settings are listed. (yes/partial/no/NA)
    \end{itemize}
\end{enumerate}
\end{document}